\documentclass[runningheads]{llncs}
\pdfoutput=1
\usepackage{graphicx, wrapfig}
\usepackage{amsfonts}
\usepackage{todonotes}
\usepackage{color}
\usepackage[font=small,labelfont=bf]{caption}
\begin{document}
\title{Microscopic fine-grained instance classification through deep attention}
\titlerunning{Microscopic fine-grained instance classification}
%
\author{Mengran Fan\inst{1}
 Tapabrata Chakraborti\inst{1} 
 Eric I-Chao Chang\inst{2}
 Yan Xu \inst{2,3} \and
 Jens Rittscher \inst{1}}
\authorrunning{M. Fan et al.}
 \institute{Institute of Biomedical Engineering, Dept. of Engg. Science, Univ. of Oxford, UK  \\
 \email{mengran.fan; tapabrata.chakraborty; jens.rittscher  \ @eng.ox.ac.uk}
\and
Microsoft Research, Beijing, China \\
 \email{echang@microsoft.com}
\and
Department of Biology and Medicine, Beihang University, Beijing, China \\
\email{xuyan04@gmail.com}}
\maketitle              
\begin{abstract}
Fine-grained classification of microscopic image data with limited samples is an open problem in computer vision and biomedical imaging. Deep learning based vision systems mostly deal with high number of low-resolution images, whereas subtle detail in biomedical images require higher resolution. To bridge this gap, we propose a simple yet effective deep network that performs two tasks simultaneously in an end-to-end manner. First, it utilises a gated attention module that can focus on multiple key instances at high resolution without extra annotations or region proposals. Second, the global structural features and local instance features are fused for final image level classification. The result is a robust but lightweight end-to-end trainable deep network that yields state-of-the-art results in two separate fine-grained multi-instance biomedical image classification tasks: a benchmark breast cancer histology dataset and our new fungi species mycology dataset. In addition, we demonstrate the interpretability of the proposed model by visualising the concordance of the learned features with clinically relevant features.

\end{abstract}
\section{Introduction}

Fine-grained image classification, which focuses on distinguishing subtle visual differences between classes, is an open problem in biomedical image analysis. Deep learning has led to a remarkable progress in fine-grained classification on large-scale natural images \cite{zheng2019looking,yang2018learning,rodriguez2018attend}. Despite the important advances in computer vision, it is usually challenging to achieve the same success on specific biomedical image classification tasks \cite{zhang2019medical,yan2019breast}. To sum up, current methods mainly face three challenges. Due to the cost of data acquisition and the limited availability of specimens, well-organised medical datasets in medical usually tend to be small, which limits the representation ability of deep networks. The main reason is that the current state-of-the-art convolutional neural networks (CNN) are capable of extracting semantically meaningful features on large-scale datasets. When training data is limited deep networks may overfit and may bias the classification result on confounding background clutter.

\begin{figure*}[t]
\includegraphics[width=\textwidth,height = 2.5cm]{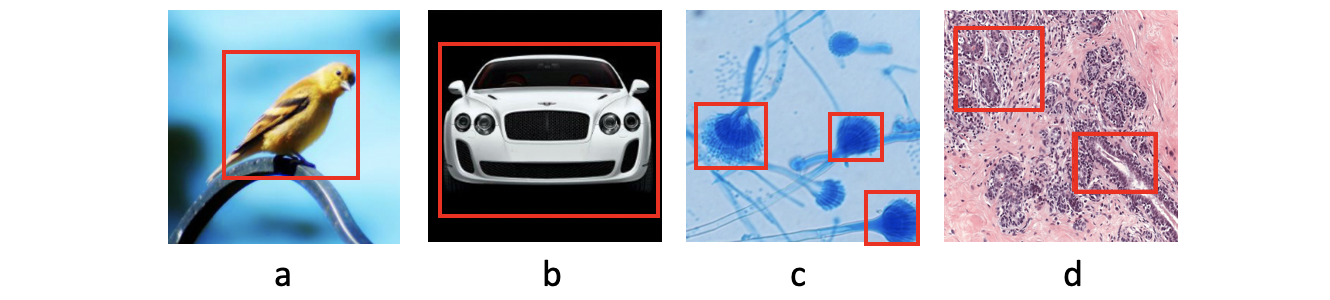}
\caption{\textbf{Classification challenges in biomedical imaging.} Compared to large-scale datasets on natural images (a-b), well-organised biomedical datasets (c-d) tend to be small and require expensive human expert annotations. Secondly, while we typically expect one centre-aligned instance (red box) in natural images, there are often multiple discriminative instances (red boxes) in biomedical images, which poses another challenge for feature learning. Finally, interpretability needs to be considered when developing a reliable medical image analysis system.}
\label{f:data-sample}
\end{figure*}

Especially when working on the microscopic scale, multiple instances (e.g. glands, vessels, or crypts) need to be considered. This seriously restricts the adaptation of existing methods in the fine-grained classification of natural images. For instance, we randomly select samples (Fig.\ref{f:data-sample} (a) - (b)) from the most popular fine-grained datasets in computer vision (CUB-200-2001 \cite{wah2011caltech} and Stanford Cars \cite{krause20133d}), where there is mostly one centre-aligned instance in an image. Although a large number of strategies have been proposed to detect the discriminative parts (e.g., head, belly for birds) in such images, the size and layout of the detected components are almost identical for each image. In comparison to natural images, biomedical images (Fig.\ref{f:data-sample} (c) - (d)) may have a wide variety of discriminative instances (regions) with different sizes and densities, leading to more complicated structural information and a larger within-class variation. This motivates the need for investigating methods for building comprehensive and discriminative feature representations that can be applied in this domain. Thirdly, apart from the accurate prediction, the interpretability also plays a crucial role in a reliable medical image classification system \cite{weese2016four}. In this work, we propose a novel attention-based classification network that is capable of jointly localising discriminative instances and enhancing consistent fine-grained feature learning in an end-to-end fashion. The main contributions of this paper are: (1) A lightweight gated attention module where the most discriminative instances can be localised simultaneously without requiring any part annotations or redundant region proposals. (2)  A multi-task learning scheme that dynamically controls the weights of member modules and enforces the network to learn consistent instance-level features. (3) Improved  interpretability of learned features when compared with features used by human experts for decision making.

\section{Methodology}
\label{sec:method}

\begin{figure*}[t]
\includegraphics[width=\textwidth, height =8cm]{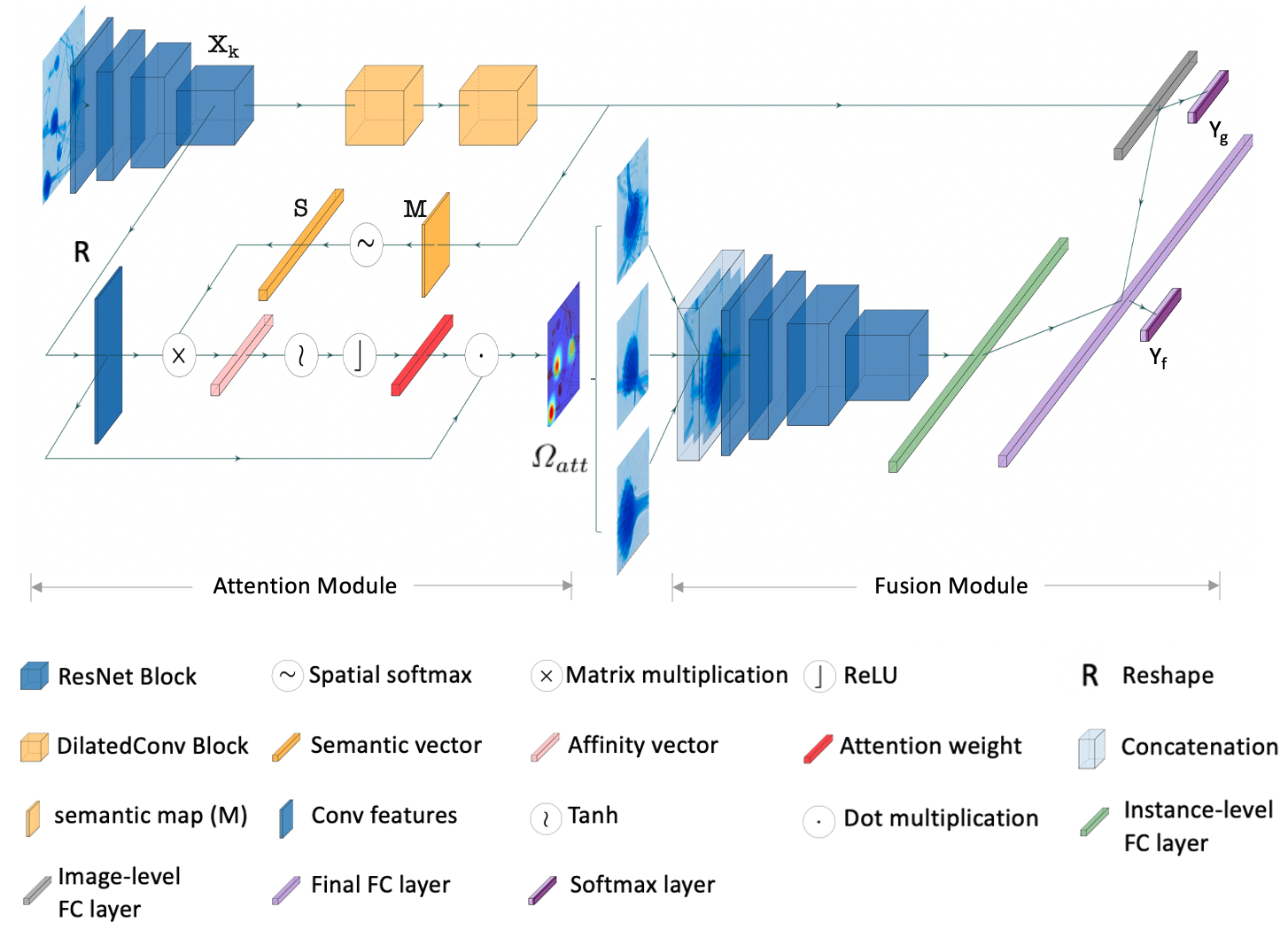}
\caption{\textbf{Framework for the proposed multiple instance fine-grained classification pipeline.}  The proposed network consists of two main modules: the attention module and the feature fusion module. The details of the attention module and definition of the variables are explained in Section \ref{ss:gated-attention}. In the fusion module, we threshold the attention map to generate a binary mask, crop these patches from the input image, resize them to a predefined size and feed them into a shared CNN model. Finally, we concatenate all instance-level features with image-level features for final prediction. The predictions in the inference stage are also conducted in this way.}
\label{f:model}
\end{figure*}
We propose a novel fine-grained multi-instance classification scheme (Fig.\ref{f:model}) that consists of two main modules: (i) a novel gated attention mechanism for discriminative instances localisation; and (ii) a feature fusion strategy that consolidates the global and local features to facilitate the final predictions.

\subsection{Gated attention module}
\label{ss:gated-attention}
In natural images, noisy background or irrelevant regions are highly variable and can be therefore naturally discarded by deep neural networks. However, for medical datasets with limited training samples, this is more difficult to achieve. Here, we propose a gated attention mechanism that is aimed to filter out the confounding channels and localise the most discriminative instances without extra part supervision or redundant region proposals. As shown in Fig.\ref{f:model}, the module first summarises a $2D$ semantic map $ S \in \mathbb{R}^ {H \times W}$ from the input convolutional feature maps $ X \in \mathbb{R}^ {C \times H \times W }$. Furthermore, we use the high-level semantic information to filter out the confounding channels based on the pairwise correspondences between each input channel and the generated semantic map, thus suppressing the irrelevant background and localising the most discriminative instances.

\textbf{Semantic Modelling.} Since spatial attention weights computed for each query position are almost the same for different tasks \cite{cao2019gcnet}, 
we extract a global spatial attention map from the input features, which are shared by all query positions within an image. For multi-level semantics understanding, we first apply two dilated convolution blocks \cite{chen2017deeplab} to the output of the feature extractor $X_{k}$. The set of multi-scale features are compressed by computing the sum of all channels $M = \sum_{k=1}^{C}W^{T}X_{k}$ where $W^{T}$ is the weights of dilated blocks. This channel compression rests on the assumption that if the region is activated on most channels, the region tends to be more discriminative and to have higher likelihood of being part of the object of interest.
The final semantic map $S$ is generated by applying a spatial softmax layer that performs the softmax operation over all feature points in the aggregated map $M$, resulting in a probability distribution that roughly indicates the regions of the most discriminative instances:
\begin{equation}
S_{i,j} =  \frac{\exp{(M_{i,j})}}{\sum_{l=1}^{H}\sum_{k=1}^{W}\exp{(M(l,k)})}
\label{equation:semantic_map}
\end{equation}

\textbf{Gated Mechanism.} To measure the discriminability of each channel, we capture the spatial correspondence scores via conducting matrix multiplication over $X^{T}$ and $S$, where $X^{T}$ is the original convolutional feature maps with the shape of $c \times hw$. For example, $X_{k}$ is the $k^{th}$ channel of the input feature maps, containing its specific semantic responses. So $X^{T}_{k}S$ is the importance coefficient that indicates the semantic representation power of this channel. Our method is different from traditional channel-based self-attention mechanisms \cite{zheng2019looking,fu2019dual} that usually directly capture the pairwise inter-channel dependencies by calculating $X^{T}X$. In order to enhance the specific semantics, we summarise a global high-level semantic map and use it as a template to quantify the representation capability of each channel by $X^{T}S$. Therefore, we apply such a mechanism to obtain the $1D$ coefficient vector $X^{T}S \in \mathbb{R}^ {C \times 1}$ rather than $X^{T}X \in \mathbb{R}^ {C \times C}$. 

To filter out the confounding channels, the hyperbolic tangent ($tanh$) and ReLU activation functions are used to normalise the discriminability coefficient among all channels. As a result, a set of gated weights is obtained, selects channels that look at the most discriminative regions. In particular, the gated weight is approximately $1$ for the most informative channels, and approximately $0$ for the channels highlighting the irrelevant background (Fig. \ref{f:gated_att}). The gated activation layer can also be regarded as a filter which enforces the model to ignore the confounding channels and pay attention to more informative channels. Consequently, to let the attention module focus on multiple instances, we model the final attention map as a gated average of the outputs of the original channels.

\begin{equation}
\Omega_{att} = \frac{1}{C}\sum^{C}_{k=1} ( X^{T}_{k} \odot ReLU (\tanh{( X^{T}_{k} \otimes S)}))
\end{equation}

\subsection{Multi-task Loss Function}
Different from traditional two-stage frameworks consisting of two separate networks, the multi-task loss aims to enable the model to jointly learn multi-instance localisation and image classification in an end-to-end fashion. Specifically, our network is optimised by a global attention loss and a final fusion loss:
\begin{equation}
\mathcal{L} = \lambda L_{\mathcal{G}}(Y_{g},Y^{\star}) + (1 - \lambda)(L_{\mathcal{F}}(Y_{f},Y^{\star}))
\end{equation}
where $L_{\mathcal{G}}$ and $L_{\mathcal{F}}$ are standard cross entropy losses with respect to the outputs of the global image-level network $Y_{g}$ and the proposed multi-instance fusion network $Y_{f}$, respectively. $Y^{\star}$ represents the ground truth label and the parameter $\lambda$ is initialised as $1$ and gradually decreased during training. As a result, the network initially focuses on extracting global image-level features, and increases the contribution of discriminative instance-level features during training. 

\section{Evaluation}
\label{sec:evaluation}
All input images were resized to $224 \times 224$, and a Resnet-18 was used to extract global image-level information from down-sampled images. After instance localisation, extracted patches were scaled to $336 \times 336$, and fed to a Resnet-50 for final image-level prediction. Other CNN architectures could be used instead. To improve training efficiency, pre-trained weights from the ImageNet dataset were used for initialisation. Mini batch size was set to $16$. We used the stochastic gradient descent (SGD) optimiser with an initial learning rate of $0.05$ that was multiplied by $0.1$ after every $50$ epochs. The initial weight score $\lambda$ in the loss function is $1$, and reduced by $0.1$ after every $20$ training epochs. The publicly available MXNet library was used to implemented the model, training was performed on two NVIDIA GeForece 1080 Ti GPUs.

\textbf{Evaluation and performance analysis on new fungi species dataset.}To the best of our knowledge, this is the first attempt for bringing deep learning based approaches to fungal species identification. $2151$ microscopy images from $59$ patients were collected in collaboration with the Peking Union Medical College Hospital. In this dataset, we particularly focus on five most common species involved in human disease: (1)Aspergillus fumigatus, (2)Aspergillus flavus, (3)Aspergillus niger, (4)Aspergillus terreus and (5)Aspergillus nidulans. We provide quantitative results and compare it with recent competing methods. We also benchmark the performance of the novel gated attention mechanism with other attention schemes. For all experiments, we randomly split the samples in each class in a ratio of $1:3$ for constructing testing and training sets. 

\textbf{Quantitative comparison with competing methods}
We evaluated the effectiveness of the proposed method by comparing it with several state-of-the-art fine-grained classification methods. All of the compared methods were trained with the same backbone network and computing environment. From the comparison shown in Table \ref{t:fungi-results}, we observe that our method achieves the best performance when compared with other fine-grained classification methods.

\begin{table}
\parbox{.45\linewidth}{
\caption{\textbf{Results on Fungi species dataset.}}
\centering
\label{t:fungi-results}
\begin{tabular}{c|c}
\hline
\textbf{Methods} & \textbf{Accuracy} \\
\hline
Resnet-50 \cite{he2016deep}  &0.907\\
Residual attention \cite{wang2017residual} &0.867\\
Attend \& Rectify\cite{rodriguez2018attend}&0.871\\
Trilinear attention \cite{zheng2019looking}& 0.883\\
NTS Network \cite{yang2018learning} & 0.914\\
\textbf{Our method} & \textbf{0.951}\\
\hline
\end{tabular}
}
\hfill
\parbox{.5\linewidth}{
\caption{\textbf{Comparison of attention mechanisms.}}
\centering 
\label{t:att}
\begin{tabular}{c|c}
\hline
\textbf{Attention Mechanisms} & \textbf{Accuracy} \\
\hline
Spatial attention \cite{wang2018non} &0.859\\
Channel-wise attention \cite{zheng2019looking}&0.883\\
Dual attention  \cite{fu2019dual} & 0.901\\
Squeeze-Excitation attention \cite{hu2018squeeze}& 0.939\\
Global Context attention \cite{cao2019gcnet}& 0.937\\
\textbf{Our Gated Attention}& \textbf{0.951}\\
\hline
\end{tabular}
}
\end{table}

\begin{figure}
\centering
\includegraphics[width=0.65\textwidth]{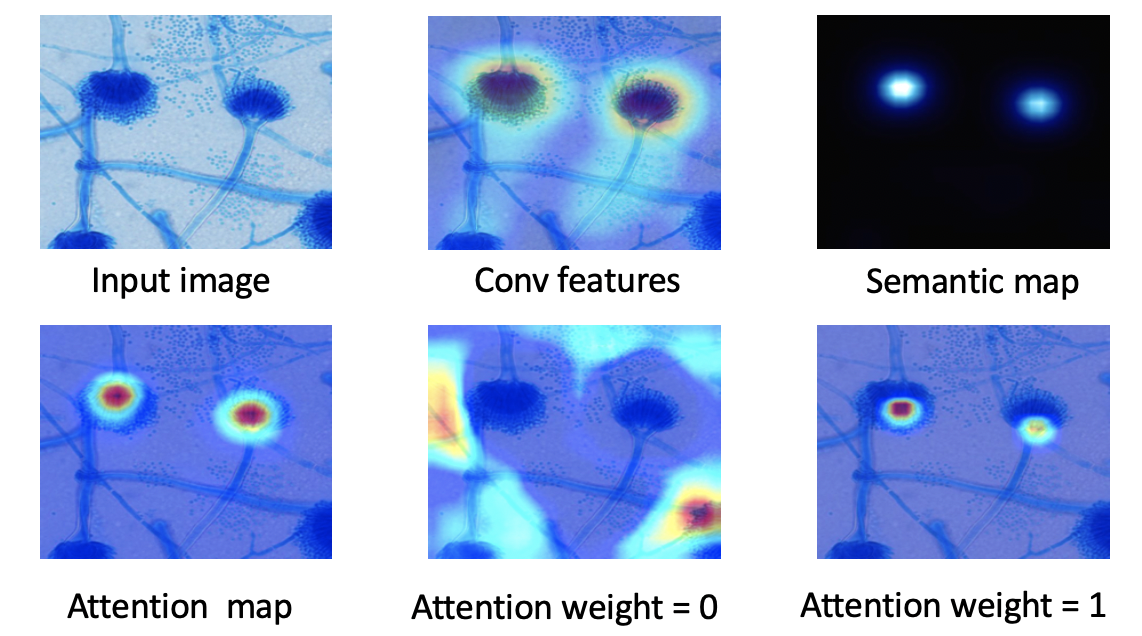}
\captionof{figure}{\textbf{Visualization of the gated attention module for one sample image.} The top row shows the input image, original convolutional feature map and the corresponding semantic map (defined in Eq.\ref{equation:semantic_map}). The bottom row shows the final attention map and two representative channels with the lowest and highest attention.  }
\label{f:gated_att}
\end{figure}

\textbf{Evaluation of gated attention mechanism.}
To measure the effectiveness of our gated attention module we compared it with other existing attention mechanisms but using the same sampling strategy, feature fusion scheme and loss function. We only modified the attention-based instance localisation module in the baseline model, and investigate the performance of different attention mechanisms. Table \ref{t:att} shows the results of integrating different attention modules in our classification framework, and Fig. \ref{f:gated_att} depicts a visualisation example of each step in our gated attention module. Our gated attention mechanism not only outperforms all other attention modules but it also suppresses the confounding information, demonstrating the effectiveness and localisation ability. 

\begin{table}[ht]
\caption{\textbf{Results on breast cancer dataset.}}
\begin{center}
\begin{tabular}{l|c|l|c}
\hline
\textbf{Methods} & \textbf{Accuracy} & \textbf{Methods} & \textbf{Accuracy}  \\
\hline
Vgg19 \cite{kohl2018assessment}& 0.925 & 
Inception-v3 \cite{kohl2018assessment}& 0.913\\
DenseNet-161 \cite{kohl2018assessment}& 0.940 &
Model Fusion \cite{rakhlin2018deep} & 0.925\\
AlexNet \cite{nawaz2018classification} & 0.813 &
ResNet-152 \cite{cao2018improve} & 0.830\\
RFSVM-All \cite{cao2018improve} & 0.930 &
Ensemble \cite{vang2018deep}& 0.825\\
Refined Ensemble \cite{vang2018deep}& 0.875 &
Two-stage network \cite{golatkar2018classification} & 0.850 \\
Hybrid deep network \cite{yan2019breast} & 0.913 &
\textbf{Our method} & \textbf{0.970}\\
\hline
\end{tabular}
\end{center}
\label{t:results-histology}
\end{table}

\textbf{Evaluation on Breast Cancer Histology images.} 
The BreAst Cancer Histology images (BACH) benchmark dataset \cite{aresta2019bach} is used to investigate the method's ability for histology images. This dataset consists of 400 high-resolution ( $ 2018 \times 1356$ ) Hematoxylin and eosin stained microscopy images, with an even distribution over four classes. Each image is labeled as one of four types: 1) normal, 2) benign, 3) in situ carcinoma and 4) invasive carcinoma, according to the predominant tissue type. We randomly perform a $75\% - 25\%$ split for training and testing. Table \ref{t:results-histology} shows the classification results on breast cancer histology images. We compared the best classification accuracy over several advanced methods in the case of the 400 images provided by the challenge organizer. Our approach achieves the best classification performance with $0.970$, showing that our network can be effectively applied to the classification tasks of histology images.

\begin{figure*}[t]
\includegraphics[width=\textwidth,height = 4.5cm]{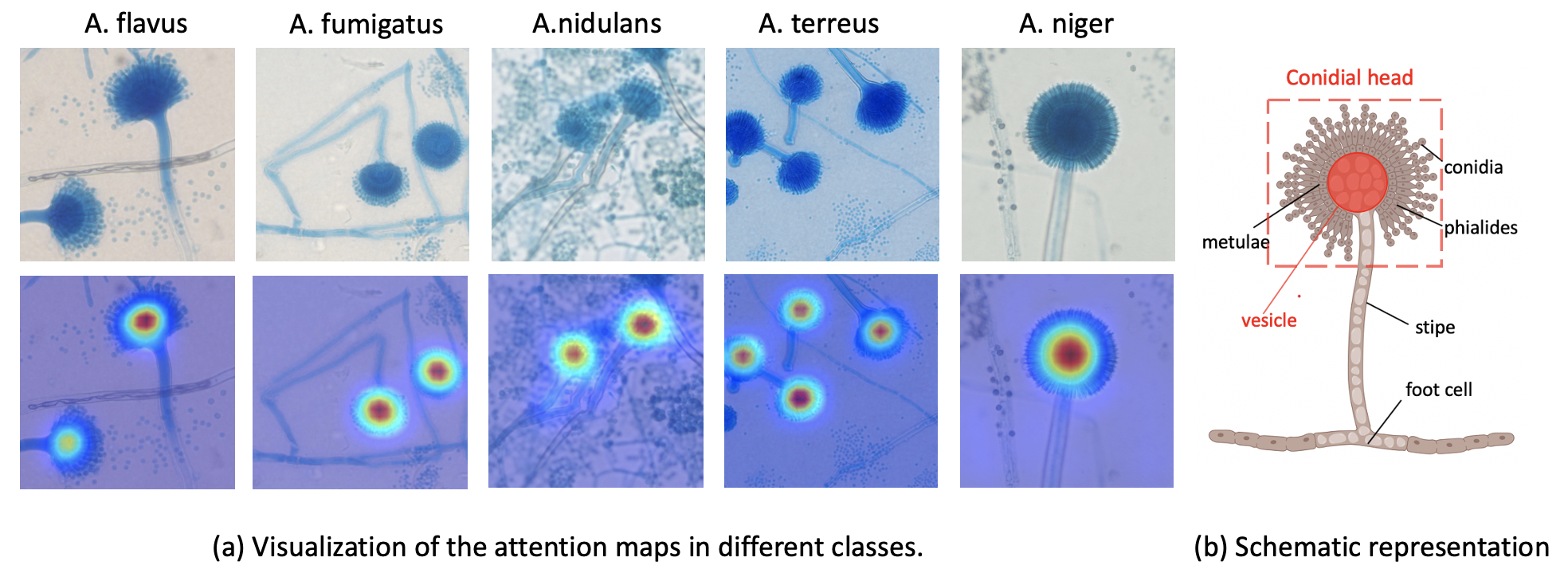}
\caption{\textbf{Clinical alignment on fungi dataset.} Clinicians mainly rely on the morphology assessment of conidial heads, especially vesicles (shown in (b)), to differentiate Aspergillus species. The attention maps (shown in (a)) generated by the proposed network consistently match the guideline for clinical decision making. \it}
\label{f:fungi}
\end{figure*}

\begin{figure*}[h!]
\includegraphics[width=\textwidth,height = 4.7cm]{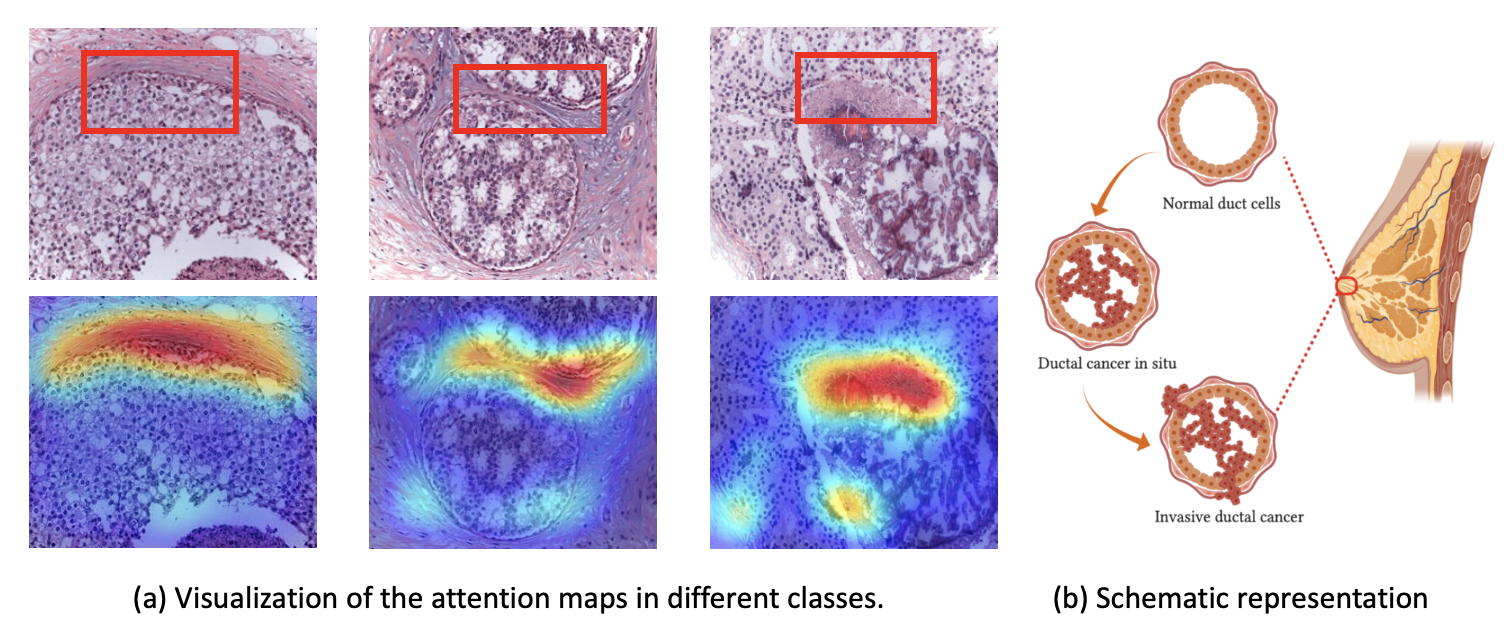}
\caption{\textbf{Clinical alignment on breast cancer dataset.} The first row in (a) shows the examples of $in$ $situ$ carcinomas with bounding box annotations. The attention maps shown in the second row consistently focus on the membrane boundaries, covering the human annotations . \it}
\label{f:breast}
\end{figure*}

\textbf{Interpretability and alignment with clinical background.} By analysing the concordance of the learned attention maps with well established visual clues used by human experts we evaluate their interpretability. The reader can easily appreciate the importance of this in addition to accuracy in results. \textbf{Fungal species:} In clinical practice, key criteria \cite{zulkifli2017morphological,diba2007identification} are a range of morphological features associated with the structure of conidial heads, especially the colour, size and shape of vesicles (Fig.\ref{f:fungi} (b)). Fig.\ref{f:fungi} (a) shows the sample images and corresponding attention maps of each specie. Our attention maps consistently highlight the relevance of these vesicle patterns. 
\textbf{Breast cancer:} 
A normal healthy breast duct is made up of layers of inner epithelial cells, outer myoepithelial cells and a basement membrane (see Fig.\ref{f:breast} (b)). In the case of $in$ $situ$ carcinomas, the proliferating cancer cells are restrained inside the basement membrane, whereas the cancer cells break out of the walls and invade the surrounding breast tissue in invasive cases. Thus, the intactness of the basement membrane is diagnostic relevance. To evaluate the effectiveness of learned features, bounding box annotations were generated by an expert pathologist on 100 test images. Overall, $72\%$ of the bounding boxes are covered by our network and selected examples of $in$ $situ$ carcinomas are shown in Fig.\ref{f:breast} (a). 

\section{Conclusion}
We present a simple yet effective end-to-end deep architecture that addresses the problem of fine-grained multi-instance classification from biomedical images at high resolution. It achieves this by first using a lightweight gated attention mechanism that detects multiple key instances and then combining the global structure and local instance features for a final image level classification. The proposed network is evaluated on a new fungi species classification dataset and a publicly available breast cancer dataset and achieves state-of-the-art performance. We also demonstrate in details the scope of our method as an interpretable model by showing the strong alignment of the learned features with well documented visual clues used by human subject matter experts.

\subsubsection{Acknowledgement.}  MF received financial support from the Arthritis Therapy Prgramme (A-TAP) funded by the Kennedy Trust. JR and TC are supported by the EPSRC SeeBiByte Programme Grant (EP/M013774/1). TC is also supported by the Oxford CRUK Cancer Centre. 
%
%
%


\end{document}